\documentclass{article}
\usepackage{arxiv}
\pdfoutput=1
\usepackage{amsmath}
\usepackage{subcaption}
\usepackage[utf8]{inputenc} 
\usepackage[T1]{fontenc}    
\usepackage{hyperref}       
\usepackage{url}            
\usepackage{booktabs}       
\usepackage{amsfonts}       
\usepackage{nicefrac}       
\usepackage{microtype}      
\usepackage{lipsum}
\usepackage{graphicx}
\graphicspath{ {./images/} }

\usepackage{geometry}
\title{Variational Autoencoder-based Neural Network Model Compression}

\author{
 Liang Cheng \\
  Department of Informatics\\
  University of Oslo\\
  \texttt{liangch@ifi.uio.no} \\
   \And
 Peiyuan Guan* \\
  Department of Informatics\\
  University of Oslo\\
  \texttt{peiyuang@ifi.uio.no} \\
  \And
  Amir Taherkordi \\
  Department of Informatics\\
  University of Oslo\\
  \texttt{amirhost@ifi.uio.no} \\
  \And
  Lei Liu \\
   Guangzhou Institute of Technology\\Xidian University\\
\texttt{tianjiaoliulei@163.com} \\
\And
 Dapeng Lan \\
  Shenyang Institute of Automation\\
  \texttt{landapeng@sia.cn} \\
}

\begin{document}
\maketitle
\begin{abstract}
Variational Autoencoders (VAEs), as a form of deep generative model, have been widely used in recent years, and shown great great peformance in a number of different domains, including image generation and anomaly detection, etc.. This paper aims to explore neural network model compression method based on VAE. The experiment uses different neural network models for MNIST recognition as compression targets, including Feedforward Neural Network (FNN), Convolutional Neural Network (CNN), Recurrent Neural Network (RNN) and Long Short-Term Memory (LSTM). These models are the most basic models in deep learning, and other more complex and advanced models are based on them or inherit their features and evolve. In the experiment, the first step is to train the models mentioned above, each trained model will have different accuracy and number of total parameters. And then the variants of parameters for each model are processed as training data in VAEs separately, and the trained VAEs are tested by the true model parameters. The experimental results show that using the latent space as a representation of the model compression can improve the compression rate compared to some traditional methods such as pruning and quantization, meanwhile the accuracy is not greatly affected using the model parameters reconstructed based on the latent space. In the future, a variety of different large-scale deep learning models will be used more widely, so exploring different ways to save time and space on saving or transferring models will become necessary, and the use of VAE in this paper can provide a basis for these further explorations.
\end{abstract}

\keywords{Deep Learning \and Variational Autoencoders (VAEs) \and Model Compression}

\section{Introduction}
In the last few years, the development of neural networks has been more rapid than ever before, with new models becoming more accurate and more complex. A significant portion of the currently popular models are based on some basic models, such as VGGNet\cite{Simonyan2014} based on CNN\cite{Saad2017} and LSTM\cite{Sepp1996} based on RNN\cite{David1987}, while some more novel models such as Generative Pre-trained Transformer (GPT)\cite{Tom2020} and Vision Transformers (ViT)\cite{Doso2021} are developed on top of Transformer\cite{Ash2017}. Transformer is not directly based on traditional neural networks, but it is still influenced by some of the basic models, such as RNN and FNN\cite{Bebis1994}, and improves on their limitations. As the model becomes increasingly larger, the amount of model parameters also increases significantly. For example, VGGNet, as a very deep convolutional neural network, has as many as 138M parameters for its entire network, and some of its variants even contain much more parameters if more convolutional layers are stacked to improve the model accuracy.

With the massive amount of model parameters, it becomes difficult to efficiently save or transfer models between different machines or nodes. Therefore, many researchers have worked on model compression over the past few years, a number of approaches have gained widespread popularity. For example, pruning is proposed in \cite{Pruning2015}, which aims to reduce the number of parameters required by pruning out as many connections between different nodes and layers in the network as possible without compromising the accuracy of the model, and quantization is introduced in \cite{Quantization2016}, it lowers the bit-width of the model’s weights and activations to reduce the computational and memory requirements of neural networks. Both methods are widely used in deep learning research, and they respectively target the quantity and quality of parameters to optimize the model size. However, since their goal is to use the compressed model directly while accuracy is not greatly affected, compression rate is easily to reach a bottleneck in order to strike a balance with accuracy, which is 9x to 13x in pruning and around 20x in quantization. Hypothetically speaking, the compressed model could be restored in some ways before it is used, so that the compression rate could be further boosted.

In this paper, we propose VAE\cite{VAE2014} as a model compression method. As a variant of Autoencoder\cite{AE2012}, it also consists encoder and decoder, the encoder can downscale the model parameters into latent space, which can be used as a representation of compressed model, and the decoder can be responsible for reconstructing to generate a complete model for use. Through a complete encoding-decoding process, the intermediate latent space as a compressed representation is expected to achieve higher compression rates than previous methods. Instead of directly testing the latest large-scale models, we used some basic neural network models in our experiments, including FNN, CNN, RNN, and LSTM. if the VAE-based compression method can perform well on these small models, it will lay a solid foundation for subsequent compression of large models.

\section{Methods}
\begin{figure}[h]
\centering
\includegraphics[height=4cm,width=13cm]{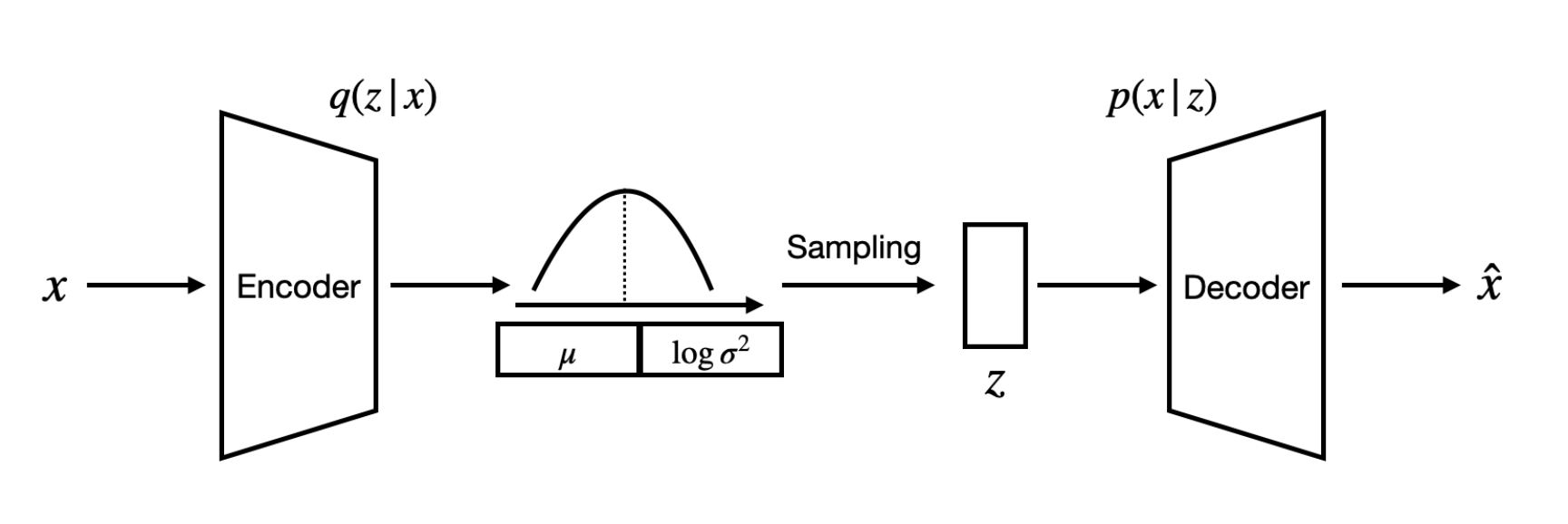}
\caption{Model Structure of VAE} \label{fig:VAE}
\end{figure}

VAE is a deep generative model improved from  Autoencoder, with modification on how to operate the latent space. The general idea behind Autoencoder is setting encoder and decoder to be neural network and using an iterative optimization process to learn the best encoding-decoding scheme. Hence, the encoder would receive input data and decoder would output the reconstructed input in each iteration, the original data and reconstructed data are compared and the errors are backpropagated through the architecture in order to update the weights of the network. Since the loss function in training Autoencoder only aims to minimize the reconstruction loss between input and output, it does not consider further for the latent space, but simply ensures that the information in the compressed representation is sufficient for decoding. As a deterministic model, Autoencoder encodes inputs as individual points in the latent space without any interpretation of uncertainty or probabilistic. And the latent space also does not have any regularization, so the structure of the latent space can be arbitrary and may not follow any meaningful distribution. In order to improve the flexibility and scalability, VAE learns probabilistic representations of data, rather than a deterministic one, and generates new samples. The model architecture of VAE is shown on Figure \ref{fig:VAE}.

VAE assumes that the observed data $\mathbf{x}$ is generated from latent variables $\mathbf{z}$, which can be described as drawing a latent variable $\mathbf{z}$ from a prior distribution $p(\mathbf{z})$ and generating data $\mathbf{x}$ from the conditional distribution $p(\mathbf{x}| \mathbf{z})$. Hence, the goal is to model the joint distribution $p(\mathbf{x}, \mathbf{z})=p(\mathbf{z}) p(\mathbf{x} | \mathbf{z})$. However, the marginal likelihood $p(\mathbf{x})=\int p(\mathbf{x}, \mathbf{z}) d \mathbf{z}$ is often complex and intractable because it requires integrating over all possible values of $\mathbf{z}$, which makes it difficult to directly optimize the likelihood $p(\mathbf{x})$. To bypass this intractability, VAE uses variational inference\cite{inference2019}. Instead of directly computing true posterior distribution $p(\mathbf{z}| \mathbf{x})$, VAE approximates it with a variational distribution $q(\mathbf{z}| \mathbf{x})$. This variational distribution can be any distribution but often as a Gaussian distribution. The idea is to make 
$q(\mathbf{z}| \mathbf{x})$ as close as possible to the true posterior distribution $p(\mathbf{z}| \mathbf{x})$.

The objective of a VAE is to maximize the Evidence Lower Bound (ELBO), which is a lower bound on the log-likelihood of the observed data. The ELBO is derived as follows:
\begin{equation}
\log p(\mathbf{x})=\log \int p(\mathbf{x}, \mathbf{z}) d \mathbf{z}=\log \int \frac{p(\mathbf{x}, \mathbf{z}) q(\mathbf{z} | \mathbf{x})}{q(\mathbf{z} | \mathbf{x})} d \mathbf{z}.
\end{equation}

By applying Jensen's inequality:
\begin{equation}
\log p(\mathbf{x}) \geq \mathbb{E}_{q(\mathbf{z} | \mathbf{x})}\left[\log \frac{p(\mathbf{x}, \mathbf{z})}{q(\mathbf{z} | \mathbf{x})}\right]=\mathbb{E}_{q(\mathbf{z} | \mathbf{x})}[\log p(\mathbf{x} | \mathbf{z})]-\mathrm{KL}(q(\mathbf{z} | \mathbf{x}) \| p(\mathbf{z})).
\end{equation}

$\mathrm{KL}(q(\mathbf{z} | \mathbf{x}) \| p(\mathbf{z})$ above is the Kullback-Leibler (KL) divergence between the approximate posterior and the prior, which acts as a regularizer. The ELBO is thus given by:
\begin{equation}
\mathrm{ELBO}=\mathbb{E}_{q(\mathbf{z} | \mathbf{x})}[\log p(\mathbf{x} | \mathbf{z})]-\mathrm{KL}(q(\mathbf{z} | \mathbf{x}) \| p(\mathbf{z})).
\label{eq:loss}
\end{equation}

Maximizing the ELBO is equivalent to minimizing the reconstruction error (first term) and the KL divergence between the approximate and true posterior (second term).

To optimize the ELBO using gradient-based methods, backpropagating through the stochastic sampling of $\mathbf{z}$ is required. The reparameterization trick enables this by expressing $\mathbf{z}$ as a deterministic function of a parameter $\boldsymbol{\mu}(\mathbf{x})$, $\boldsymbol{\sigma}(\mathbf{x})$, and a random variable $\boldsymbol{\epsilon}$ drawn from a simple distribution (like a standard normal distribution):
\begin{equation}
\mathbf{z}=\boldsymbol{\mu}(\mathbf{x})+\boldsymbol{\sigma}(\mathbf{x}) \odot \boldsymbol{\epsilon}, \quad \text { where } \quad \boldsymbol{\epsilon} \sim \mathcal{N}(0, \mathbf{I}).
\end{equation}

This reparameterization allows gradients to be propagated through $\boldsymbol{\mu}$ and $\boldsymbol{\sigma}$, enabling efficient optimization.

\section{Data}
There are two different kinds of data used in this project. The first is MNIST dataset, which is a handwritten digits database has a training set of 60,000 examples and a test set of 10,000 examples. This dataset is used to train the basic models used for the model compression experiment, including FNN, CNN, RNN, and LSTM. The parameter sets from these four trained neural networks are the data used for training VAE, which is also the model compression experiment. Even though these basic models are not large models, it can still be time-consuming to train enough of them and thus have enough model parameters VAE training. Training data and validation data are generated from these true model parameters by randomly adding noise to the different positions. For each set of true model parameters, 80 test data and 20 validation data are generated, while the true model parameters are used as test data.

The size of the parameter set varies from model to model. Although each model is used to recognize handwritten digits and the model structure is not complex, the number of parameters for each model varies due to the overall network structure as well as the layers used are different in the different networks. The information about neuron and parameter of each model is shown on Table \ref{tab:info}.

\begin{table}[h]
\centering
\small
\caption{Models information.}
\begin{tabular}{l|c|c}
& Neurons & Number of Parameters \\
\hline \# FNN & {$[784,200,100,60,30,10]$} & 185,300 \\
\hline \# CNN & {$[784,3136,1568,588,200,10]$} & 122,270  \\
\hline \# RNN & {$[28,128,128,10]$} & 54,538  \\
\hline \# LSTM & {$[28,128,128,10]$} & 214,282  \\
\end{tabular}\label{tab:info}
\end{table}

\section{Experiments \& Results}
\subsection{Data Preprocessing}
As mentioned above, the size of the parameter set for each model is different, and it's also relatively large compared to the size of most data commonly used with VAE (e.g., images or time series), so we trained VAE for each neural network separately. The first step in data preprocessing is to put the parameters of the different layers in the model into a one-dimensional sequence. Since the number of parameters contained in each layer of the neural network may vary, flattening the parameter set allows the VAE to process more easily. After getting the flattened data, the next step is to chunk the data. Since the one-dimensional sequence contains all the parameters of a model, and normal neural networks cannot process such a long sequence at once in the input layer, chunking can avoid a crash in model training. In this experiment, we set the length of the chunks to be 2048. For each parameter set where the last chunk is not enough for 2048, we use 0 to fill it and record the padding length, so that we can remove the values on padding positions in the reconstructed output from VAE.

\subsection{Experiments}
In this research, we conducted two aspects of the experiment, first exploring the degree of model compression using CNNs, and then exploring the usability of this compression method using different neural network models. Although different data were used in the experiment, the environmental settings of the experiment as well as the training conditions of the VAE were maintained. All model training and tests were performed on a MacOS 14.3 platform consisting of the Apple M3 Pro CPU and 18 GB RAM. The code was implemented in PyTorch. The loss function used in VAE training is introduce in Equation \ref{eq:loss}. The total number of parameter set of each neural network used in this work is 101, of which 1 real parameter set was used as test data and the remaining 100 parameter sets were generated based on the real parameters and were divided into training and validation sets in a ratio of 4:1. A limit of 500 epochs was set for training, subject to early stopping to avoid model overfitting. 

In the first part of experiment, we began with compression rate around 20x, which downscales the chuck size from 2048 to 128. Then we gradually reduced the size of the latent space from 128 to 64 in each VAE training, and the test accuracy almost kept consistent during the whole process. Therefore, we decided to use 64 as the size of latent space in the second part of experiment. Next, we extract the parameter sets from the trained FNN, CNN, RNN, and LSTM for training VAE, and compare their performance using the original parameters as well as those reconstructed through VAE in the end.

\subsection{Results}
\begin{figure*}[htbp]
    \subfloat[FNN]{%
        \includegraphics[height=5cm,width=7cm]{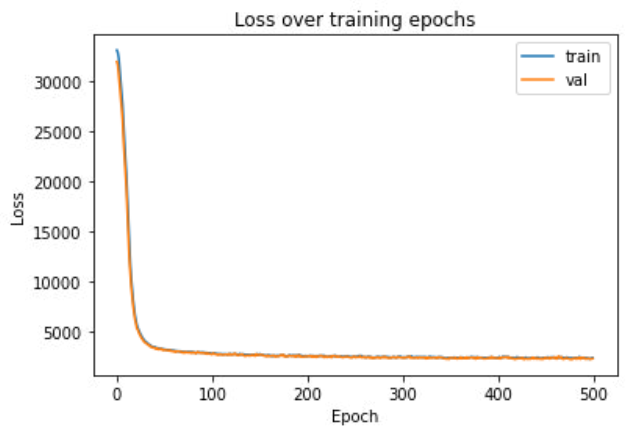}%
        \label{subfig:FNN_train}%
    }\hspace{0.4cm}
    \subfloat[CNN]{%
        \includegraphics[height=5cm,width=7cm]{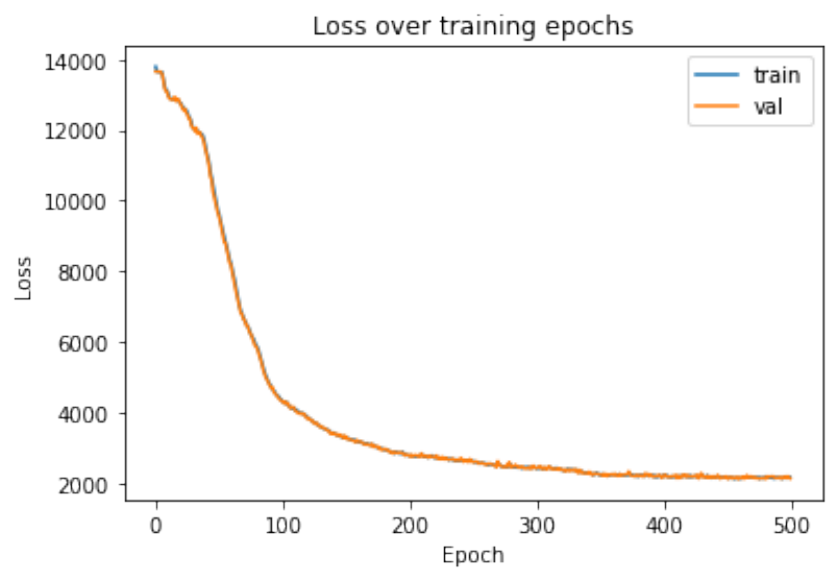}%
        \label{subfig:CNN_train}%
    }\\
    \subfloat[RNN]{%
        \includegraphics[height=5cm,width=7cm]{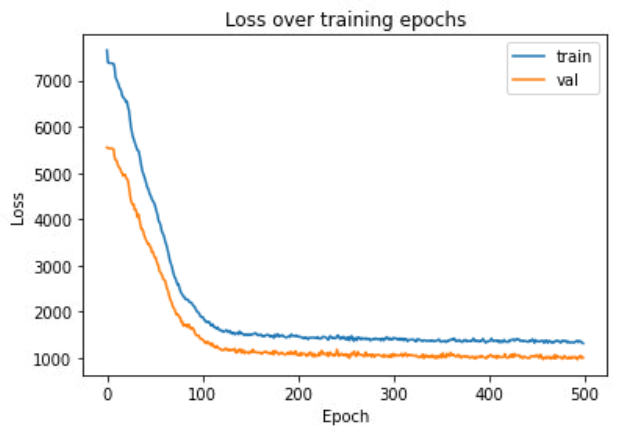}%
        \label{subfig:RNN_train}%
    }\hspace{0.4cm}
    \subfloat[LSTM]{%
        \includegraphics[height=5cm,width=7cm]{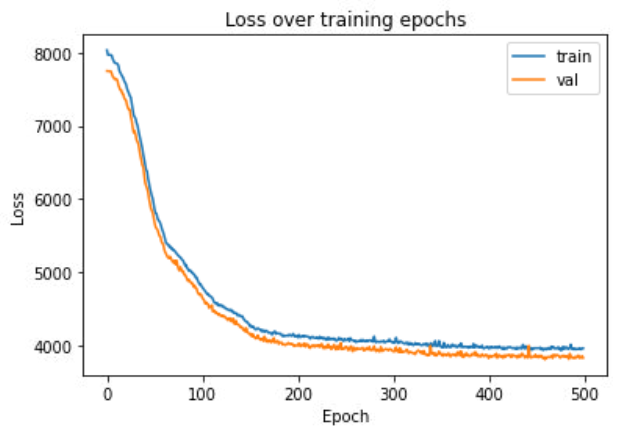}%
        \label{subfig:LSTM_train}%
    }
    \caption{VAE training process using parameters from FNN, CNN, RNN and LSTM}
    \label{fig:train}
\end{figure*}

The training process of VAEs using input data from different neural networks is shown on Figure \ref{fig:train}. As can be seen from it, there is no overfitting of the model within 500 epochs of training, even though the efficiency of learning slows down a lot after 100 to 150 epochs of training. Moreover, the learning process is not the same in the case of using different model parameters as inputs. The parameter sets from different models have values in the range of [-1, 1], and the biggest difference between them is the size. However, when using the parameters from FNN as input, the VAE clearly converges faster than other cases, but the FNN parameter set is the second largest of the four different parameter sets. This may indicate that the size of the parameter set is not the main factor affecting the learning process in VAE, but the intrinsic connection or parameter distribution of the different parameters within the parameter set is also something that VAE learns when compressing data. FNN uses only fully connected layers, and the neuron operations controlled by its parameters will be simpler than other layers such as convolutional layer, recurrent layer, etc., which possibly make its parameters easier to be learned by VAE.

\begin{figure}[htbp]
    \centering   
    \subfloat[Comparison of parameter number between different models]{
        \includegraphics[height=6cm,width=8cm]{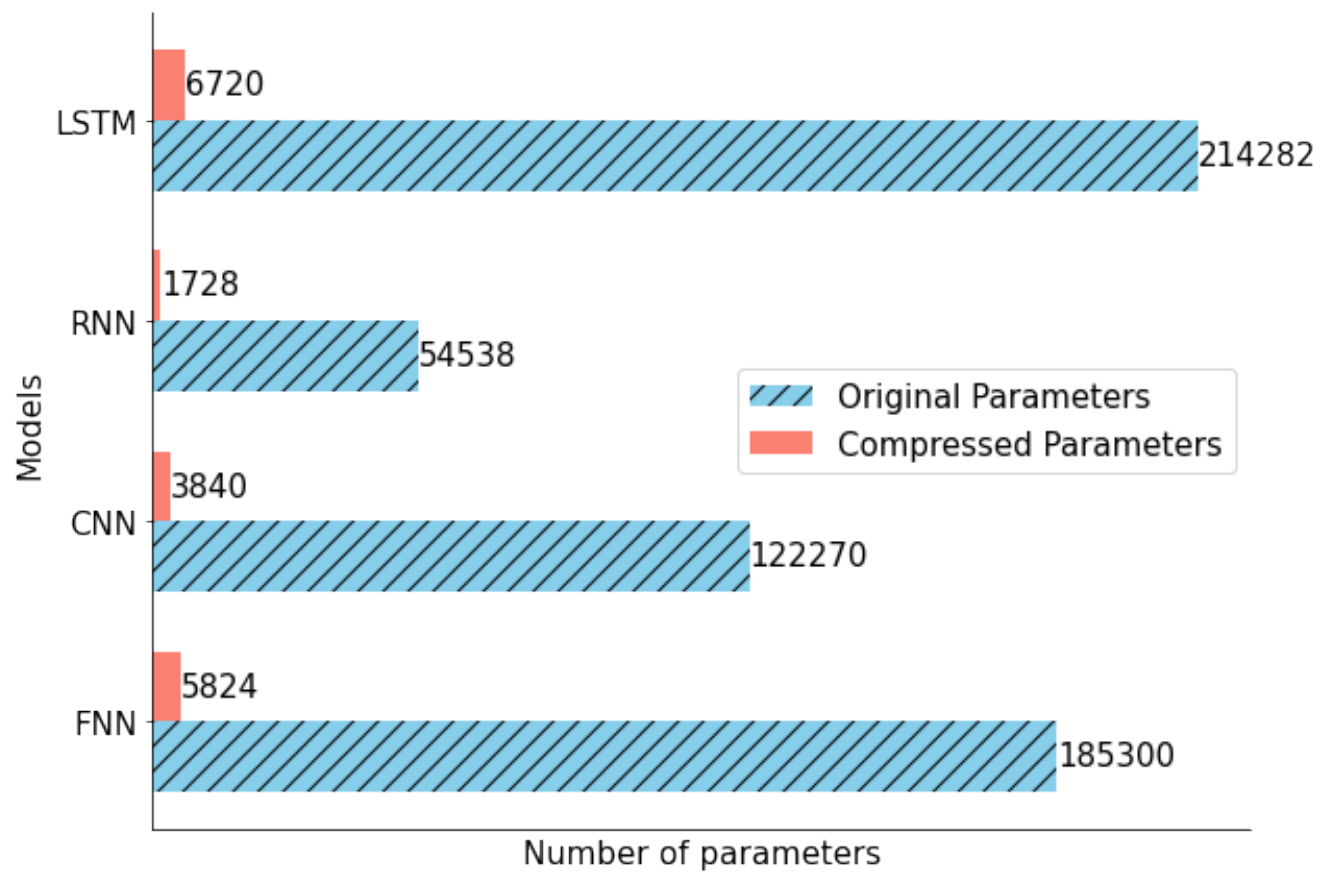}
        \label{fig:para}
    }
    \hfill
    \subfloat[VAE training time for different models]{
        \includegraphics[height=6cm,width=8cm]{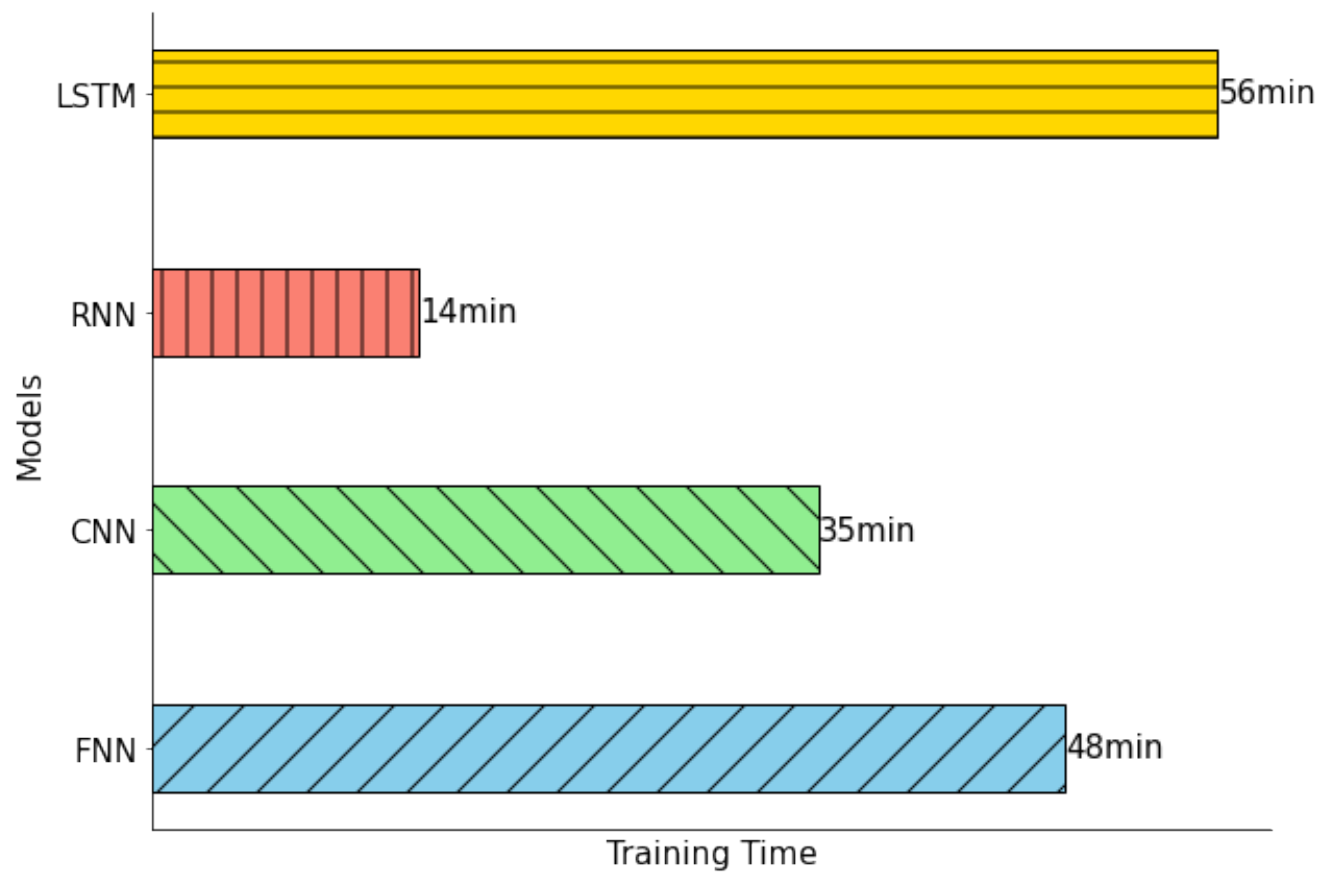}
        \label{fig:time}
    }
    \caption{Comparison of parameter size and training time across different neural networks}
    \label{fig:comparison}
\end{figure}

Figure \ref{fig:comparison} illustrates a comparison of the number of parameters included in different neural networks and their corresponding training times for the VAE. Although VAE's encoder is changing the length of each one-dimensional input sequence from 2048 to 64, each model can achieve a compression rate of slightly more than 30x, but the compression rates are not identical. This is because the size of each parameter set is different, resulting in a different number of 0 patches added in the padding portion of the data preprocessing. By comparing Figure \ref{fig:para} and Figure \ref{fig:time}, it can be seen that the ratio between the training times of different VAEs is the same as the ratio between their corresponding input sizes, even though their learning processes and learning efficiencies are not the same. A better understanding of how the size and content of the parameter set affects the training time as well as the process will play an important role in optimizing the compression of the large models, which will be much more complex than the basic model used in this experiment.

\begin{figure}[htbp]
    \centering   
    \subfloat[Test results for CNN]{
        \includegraphics[height=6cm,width=6cm]{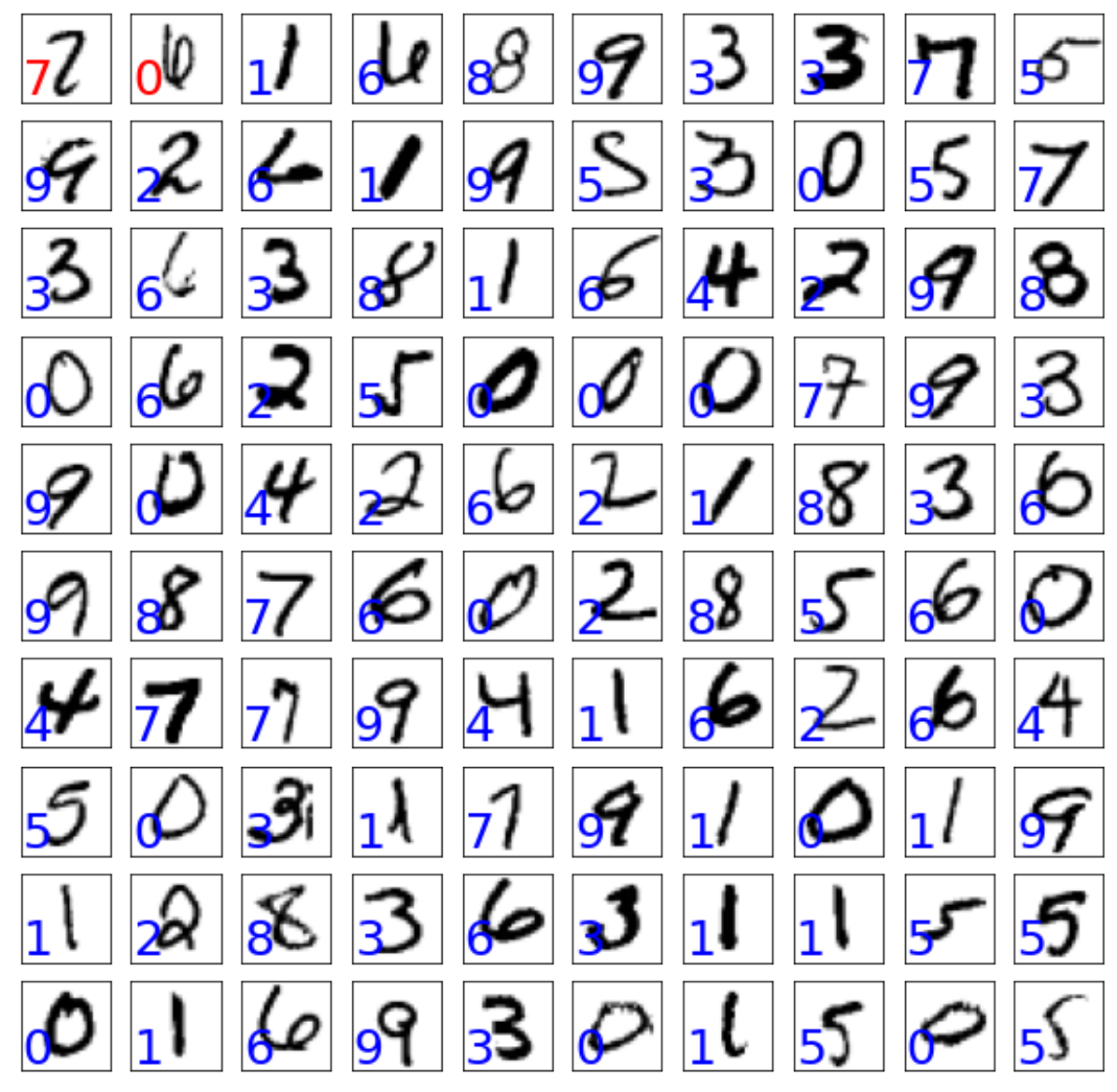}
        \label{fig:CNN1}
    }
    \hspace{0.4cm}
    \subfloat[Test results for reconstructed CNN]{
        \includegraphics[height=6cm,width=6cm]{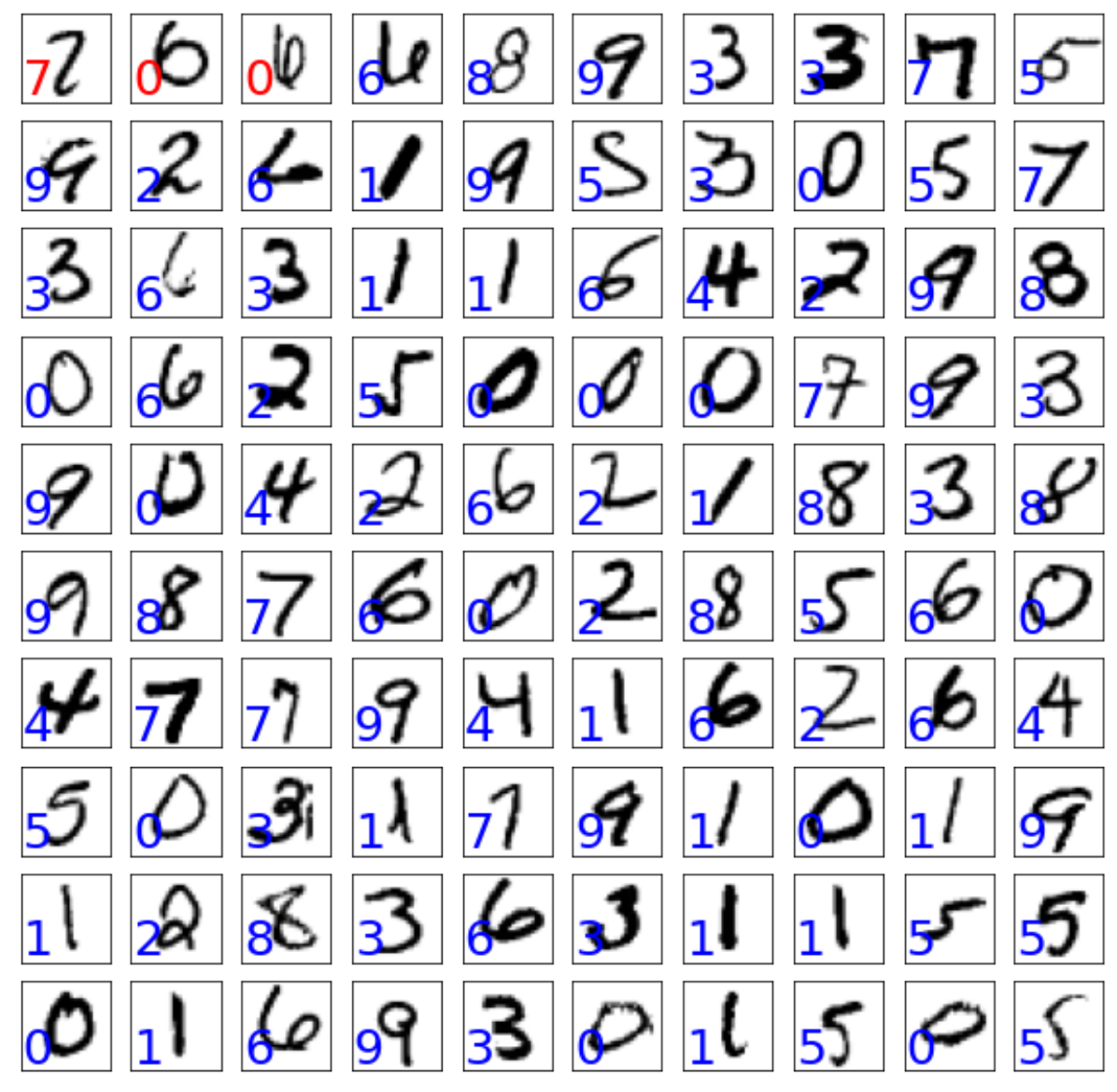}
        \label{fig:CNN_new}
    }
    \caption{MNIST test samples using CNN and reconstructed CNN from VAE}
    \label{fig:CNN}
\end{figure}

\begin{figure}[htbp]
    \centering   
    \subfloat[Test results for RNN]{
        \includegraphics[height=6cm,width=6cm]{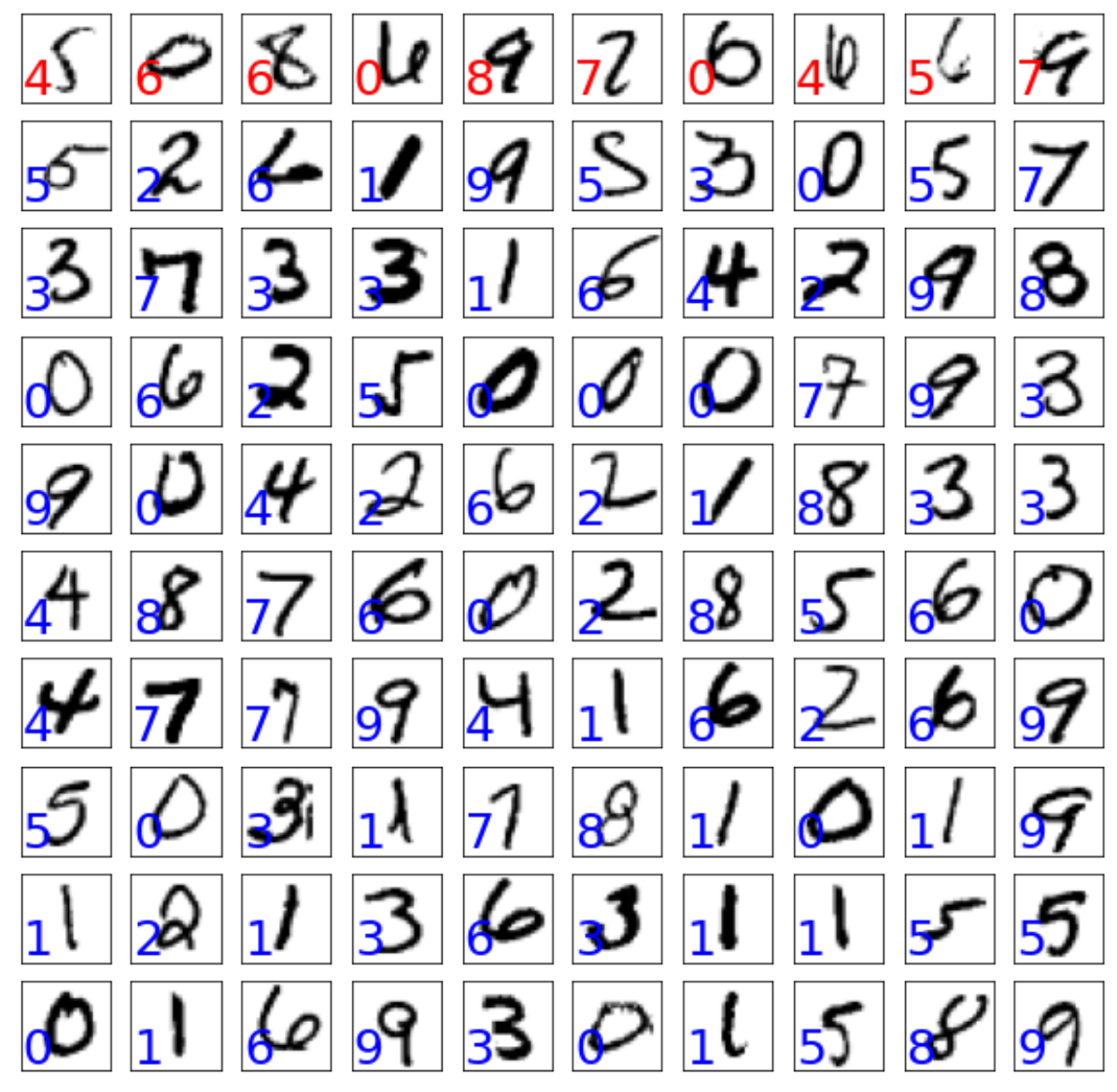}
        \label{fig:RNN1}
    }
    \hspace{0.4cm}
    \subfloat[Test results for reconstructed RNN]{
        \includegraphics[height=6cm,width=6cm]{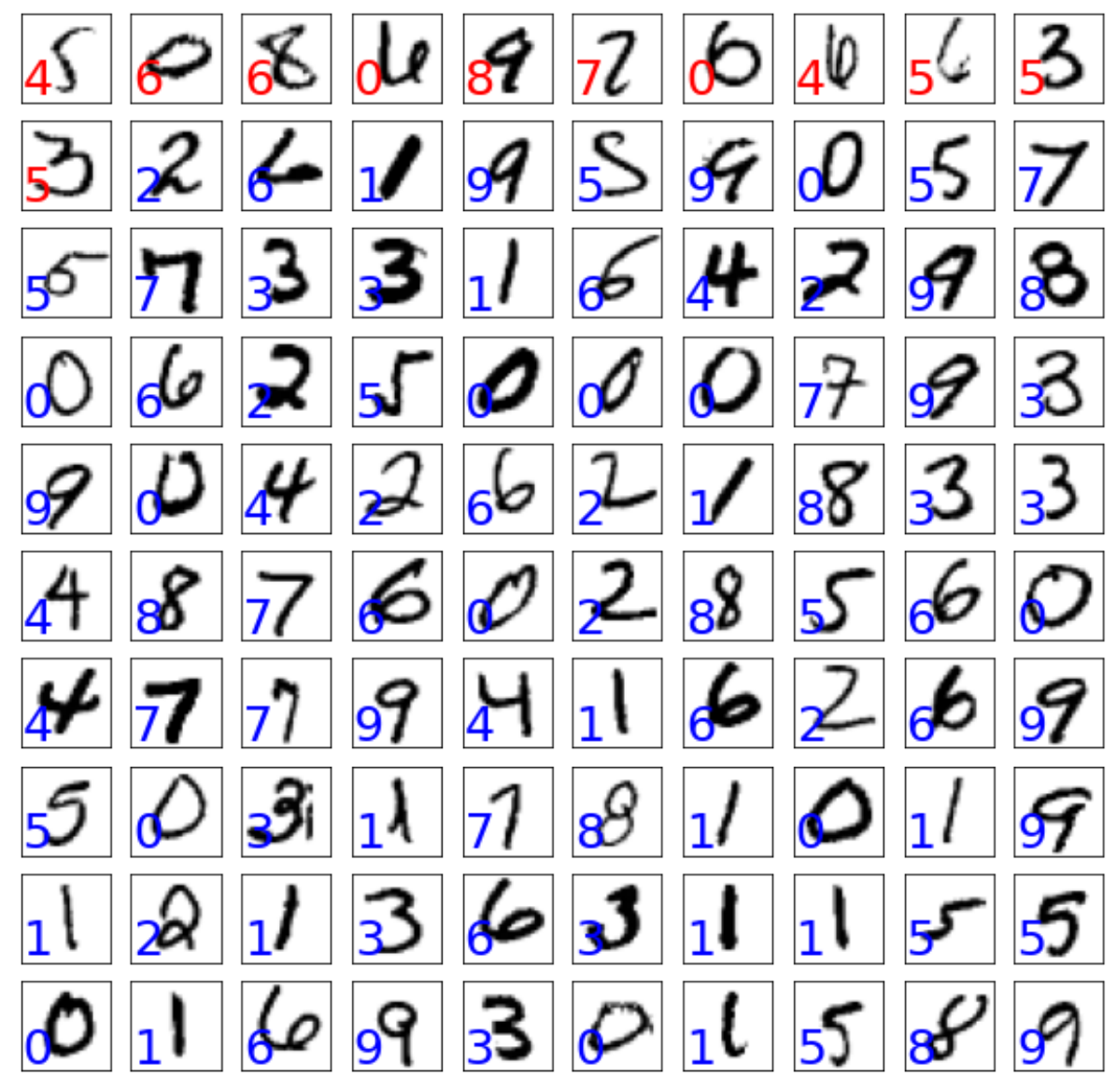}
        \label{fig:RNN_new}
    }
    \caption{MNIST test samples using RNN and reconstructed RNN from VAE}
    \label{fig:RNN}
\end{figure}

Despite the fact that the parameter set was compressed to about one-thirtieth of its original size, these reconstructed parameter sets did not significantly affect the accuracy of the respective models after reconstruction through VAE's decoder. As it can be observed from Table \ref{tab:accuracy}, the accuracy of each neural network on the MNIST test set after using the new parameter set is not much different from that when using the original parameter set, with an accuracy loss of about $1\%$ regardless of what the original accuracy was. Figure \ref{fig:CNN} and \ref{fig:RNN} show detailed test samples on MNIST for CNN and RNN using the original and new parameters, respectively. The accuracy of CNN is $98\%$ when uses the original parameters, while the RNN is slightly lower at $90\%$, and both adding only one false prediction after using the new parameters. It can be inferred that the original accuracy does not significantly impact the VAE's learning for compression and reconstruction.

\begin{table}[htbp]
\centering
\caption{Test accuracy for different models}
\begin{tabular}{l|c|c|c|c} 
& $F N N$ & $C N N$ & $R N N$ & $L S T M$ \\
\hline Accuracy of original model & $98 \%$ & $98 \%$ & $90 \%$ & $98 \%$ \\
\hline Accuracy of reconstructed model(latent space=64) & $97 \%$ & $97 \%$ & $89 \%$ & $98 \%$ 
\end{tabular}
\label{tab:accuracy}
\end{table}

\section{Conclusion \& Future Work}
Based on the size comparison of the latent space and the input in VAE, we can see that the use of VAE is able to compress the size of the neural network parameter set to one-thirtieth of the original one, and the compression rate is higher than many currently popular methods. And in the MNIST test, the model using the original parameters and the model using the parameters reconstructed from latent space through VAE have basically the same accuracy. This indicates that the 30x compression rate is not the bottleneck of the method, and that there is still room for improvement in the compression rate if the accuracy is slightly sacrificed or even if the current accuracy is continued to be preserved. These optimistic results suggest that VAE-based model compression methods are worthy of more in-depth exploration in the future.

Further research on the method can be done in two directions, namely the compression of large models and the exploration of the model structure of VAE itself. The models used in this research are all basic models, but the layers they use are similar to those in some of the large models in some aspects, such as parameter distribution and computational principles, so the compression of the large models should also be able to achieve good results, but the corresponding VAE model structure should also become more complex, because the number of parameters in the large models will be much more than that in the basic models. In the comparison of the training process, we can also see that the convergence speed of VAE is not the same for using data from neural networks with different layers, which could be related to the different parameter distributions in different neural networks, and it is also worthy of further study. As for the VAE itself, only the fully connected layer and the most basic loss function in VAE are used in this experiment. Therefore, trying different types or more complex VAEs will also help to improve the compression efficiency.

\bibliographystyle{unsrt}  
\bibliography{references}  

\end{document}